\documentclass[letterpaper]{article}
\usepackage{aaai2026}  
\usepackage{times}  
\usepackage{helvet}  
\usepackage{courier}  
\usepackage[hyphens]{url}  
\usepackage{graphicx} 
\usepackage{natbib}  
\usepackage{caption} 
\usepackage{algorithm}
\usepackage{algpseudocode}
\usepackage{amsmath,amssymb}
\usepackage{xcolor}
\usepackage{bm}
\usepackage{soul}
\usepackage{subcaption}
\usepackage{booktabs}
\usepackage{colortbl}
\usepackage{enumitem}
\definecolor{111}{rgb}{0.035, 0.53, 0.257}
\nocopyright

\providecommand{\Description}[1]{}

\title{When Good Equations Get Bad Scores: Improving Symbolic Regression Through Better Parameter Optimization}

\author{
    Boxiao Wang\textsuperscript{\rm 1},
    Kai Li\textsuperscript{\rm 1}\thanks{Corresponding author.},
    Zhiwei Chen\textsuperscript{\rm 1},
    Yang Huang\textsuperscript{\rm 1},\\
    \textbf{Runxiang Wang\textsuperscript{\rm 1},
    Ziwen Zhang\textsuperscript{\rm 1},
    Yifan Zhang\textsuperscript{\rm 1},
    Jian Cheng\textsuperscript{\rm 1}}
}

\affiliations{
    \textsuperscript{\rm 1}Institute of Automation, Chinese Academy of Sciences
}

\begin{document}
\maketitle

\begin{abstract}
Symbolic Regression (SR) plays a central role in scientific knowledge discovery by distilling mathematical equations from data.
Most existing SR methods function within a bi-level optimization framework: an outer loop searching for the discrete equation structure, and an inner loop optimizing the continuous parameters of that structure.
Crucially, parameter-fitting quality directly determines a structure's score, which in turn guides the outer-loop search. 
However, standard optimizers (e.g., BFGS) widely adopted in current systems struggle with the highly non-convex landscapes induced by nonlinear operators, often yielding poor local minima. This triggers a ``Good Structure, Bad Score'' bottleneck: correctly discovered structures are repeatedly discarded due to underestimated scores, severely degrading search efficiency.
To resolve this, we propose \textbf{SAGE-Fit} (\textbf{S}tructure-\textbf{A}ware and Semantics-\textbf{G}uided \textbf{E}valuator), an SR-native, plug-and-play optimizer that can be seamlessly integrated into existing SR algorithms. Rather than treating candidate equations as black boxes, SAGE-Fit explicitly exploits their structural and semantic priors. First, it parses the equation's algebraic structure to decouple conditionally linear parameters, dramatically collapsing the search space dimensionality. Second, it escapes local-minima traps by introducing a novel initialization strategy that enforces diversity in the semantic space rather than the deceptive parameter space. Extensive experiments demonstrate that SAGE-Fit consistently enhances evaluation fidelity and universally improves the performance of various SR systems, proving that better parameter optimization is a critical catalyst for robust symbolic discovery.
\end{abstract}

\section{Introduction}

Symbolic Regression (SR) plays a central role in AI-driven scientific discovery by distilling empirical data into interpretable mathematical equations that uncover fundamental physical laws. Because the search space of possible equations is inherently vast, modern SR methodologies—spanning Genetic Programming~\cite{koza1994genetic, schmidt2009distilling, virgolin2019linear}, Reinforcement Learning~\cite{petersen2021deep,landajuela2022unified, hayes2025deep, pmlr-v235-li24ap}, and Large Language Models~\cite{shojaee2025llmsr,merler-etal-2024-context,hua2025finetuninglargelanguagemodel,grayeli2024symbolic}—universally adopt a bi-level optimization framework to render this search tractable. In this paradigm, the discovery process is explicitly decoupled into discrete structural exploration and continuous numerical refinement. The outer loop acts as a structural architect, proposing discrete equation skeletons that define the equation's structure (i.e., mathematical operators and variables) while leaving ``placeholders'' for unknown numerical constants. The inner loop then functions as a numerical solver, filling these placeholders by optimizing continuous parameters (e.g., physical constants and scaling coefficients) to minimize the empirical error on the dataset. Crucially, this forms a tightly coupled feedback loop: the minimized loss achieved by the inner loop is returned to the outer loop as a fitness score, serving as the definitive signal that guides its subsequent structural exploration.

However, a critical asymmetry exists in current SR research. While the outer loop has benefited from sophisticated architectural innovations and powerful generative models, the inner loop still largely relies on generic optimizers such as BFGS or Nelder-Mead~\cite{udrescu2020ai,dos2024benchmarking}. This reliance assumes that off-the-shelf solvers can reliably converge to optimal parameters and accurately assess a structure's quality. In reality, physical equations typically contain nested nonlinear operators, such as exponentials and trigonometric functions, that induce highly non-convex loss landscapes, which generic continuous optimizers struggle to navigate. Consequently, structurally correct candidate equations frequently receive artificially poor fitness scores because the local optimizer gets trapped in suboptimal minima.

This phenomenon, which we term the ``\textbf{Good Structure, Bad Score}'' bottleneck, has severe repercussions for the entire discovery pipeline. Fundamentally, it triggers a high rate of optimizer-induced false negatives. When a generic solver prematurely terminates in a suboptimal region, it assigns an unjustifiably poor fitness score to an otherwise correct or highly promising structure. To quantify this severity, our empirical analysis across diverse benchmarks reveals that nearly a quarter of discarded candidates would actually outperform their corresponding incumbents after high-fidelity refitting, with this proportion reaching nearly two-thirds in extreme cases. By severely distorting the feedback signal, this failure actively penalizes the outer-loop algorithm for proposing the correct mathematical form, leading it to erroneously prune the true equation from its search space. We argue that this exposes a critical, yet largely overlooked, bottleneck in modern SR. While discovering valid mathematical structures remains a formidable challenge, the advanced capabilities of modern outer-loop algorithms are fundamentally throttled if we cannot accurately evaluate the structures they propose. Therefore, ensuring high-fidelity evaluation is an indispensable requirement for the correct equations to survive the selection process.

Resolving this requires transitioning from generic solvers to SR-native evaluators that exploit the inherent structure of mathematical equations. We propose SAGE-Fit (\textbf{S}tructure-\textbf{A}ware and Semantics-\textbf{G}uided \textbf{E}valuator), a plug-and-play parameter optimizer custom-built for SR frameworks by explicitly exploiting the dual native priors of symbolic equations. \textbf{First}, targeting the structural prior, SAGE-Fit recognizes that not all parameters are created equal, it analytically parses the equation to decouple conditionally linear parameters from non-linear ones, reducing the search space dimensionality and eliminating the spurious coupling that renders the optimization landscape highly ill-conditioned. \textbf{Second}, to target the semantic prior, we address a critical flaw in random parameter-space initialization—where multiple starting points repeatedly fall into the same suboptimal local minima—by introducing semantics-guided initialization. By evaluating the actual input-output behavioral mappings of candidates, we design a function-space farthest-point sampling mechanism to guarantee that the starting points are maximally diverse in their true semantic behavior. \textbf{Finally}, operating on the reduced non-linear manifold from these high-quality semantic seeds, SAGE-Fit employs a specialized Projected Gauss-Newton method to achieve rapid, curvature-aware local convergence.

To validate the effectiveness of our approach, we seamlessly integrate
SAGE-Fit into diverse representative SR frameworks, including
PySR~\cite{cranmer2023interpretablemachinelearningscience},
uDSR~\cite{landajuela2022unified},
LaSR~\cite{grayeli2024symbolic}, and
LLM-SR~\cite{shojaee2025llmsr}.
Extensive experiments on the LLM-SRBench
suite~\cite{shojaee2025llmsrbenchnewbenchmarkscientific}
demonstrate that upgrading the inner-loop evaluator consistently
improves both symbolic accuracy and numerical fidelity across all
base algorithms without requiring any changes to their outer-loop
search logic. These results indicate that robust inner-loop evaluation
is a fundamental prerequisite for unlocking the full discovery
potential of modern SR systems.

\section{Preliminary}
\label{pre}

\subsection{Symbolic Regression Task}
In SR~\cite{makke2024interpretable}, the learning task typically starts with a dataset
consisting of input--output pairs:
\begin{equation}
\mathcal{D}=\{(\boldsymbol{x}_i,y_i)\}_{i=1}^{n},\quad
\bm{x}_i\in\mathbb{R}^{d_x},\; y_i\in\mathbb{R},
\end{equation}
where $\bm{x}_i$ denotes a $d_x$-dimensional input vector and $y_i$ is the
corresponding scalar output. The goal is to discover an explicit analytic
equation $f(\cdot)$ such that the predicted outputs $\hat{y}_i=f(\bm{x}_i)$ accurately
approximate the ground-truth targets $y_i$, while maintaining robust generalization to unseen inputs.

\subsection{Equation Structures and Parameter Fitting}
\paragraph{Equation structures.}
In many SR methods, a candidate equation is commonly viewed as being determined jointly by a \emph{discrete symbolic structure} and \emph{continuous numerical parameters}.
The former specifies the operator symbols, variable symbols, and their combinatorial arrangement (e.g., an expression tree or a computational graph), while the latter comprises numerical quantities that must be estimated from data (e.g., constants, coefficients, frequencies, and phases).
Motivated by this widely adopted framework, we define an \emph{equation structure} as follows: all
numerically unspecified components in an expression are abstracted as
\emph{parameter placeholders}, while the discrete symbolic structure is kept unchanged.
For example, $\sin(2x)$ and $\sin(3.3x+\pi)$ can be regarded as instances of the
same structure $\sin(ax+b)$, where $a$ and $b$ are parameters to be fitted.

Formally, a structure induces a parametric function family
\begin{equation}
f(\cdot\ ;\theta):\mathcal{X}\to\mathbb{R},\qquad
\theta\in\Theta_f\subset\mathbb{R}^{d_f},
\end{equation}
where $\mathcal{X}\subseteq\mathbb{R}^{d_x}$ denotes the input space,
$\Theta_f$ is the parameter space associated with the structure $f$,
and $d_f$ is the parameter dimension (i.e., the number of parameter slots).

\paragraph{Parameter fitting.}
Given a fixed structure $f$, parameter fitting aims to estimate optimal
parameters from the dataset $\mathcal{D}=\{(\bm{x}_i,y_i)\}_{i=1}^{n}$.
We typically define the empirical loss in terms of the mean squared error (MSE):
\begin{equation}
L_f(\theta;\mathcal{D})=\frac{1}{n}\sum_{i=1}^{n}\big(y_i-f(\bm{x}_i;\theta)\big)^2.
\label{eq:mse}
\end{equation}
The parameter fitting problem for the structure $f$ can then be formulated as
\begin{equation}
\theta_f^*(\mathcal{D})\in\arg\min_{\theta\in\Theta_f} L_f(\theta;\mathcal{D}).
\end{equation}
It is worth noting that, unlike linear models of the form
``basis functions + coefficients'', 
SR often involves parameters that appear inside nonlinear operators  (e.g., $\sin(\theta x)$, $x^{\theta}$,
$\exp(\theta x)$),
which renders $L_f(\theta;\mathcal{D})$ a non-convex objective.

\begin{figure*}[t]
    \centering
    \captionsetup[subfigure]{labelformat=parens,labelsep=space}
    \renewcommand{\thesubfigure}{\Alph{subfigure}}

    \begin{subfigure}[t]{0.49\textwidth}
        \centering
        \setlength{\tabcolsep}{1.5pt}
        \renewcommand{\arraystretch}{0.9}
        \begin{tabular}{@{}cc@{}}
            \includegraphics[width=0.484\linewidth]{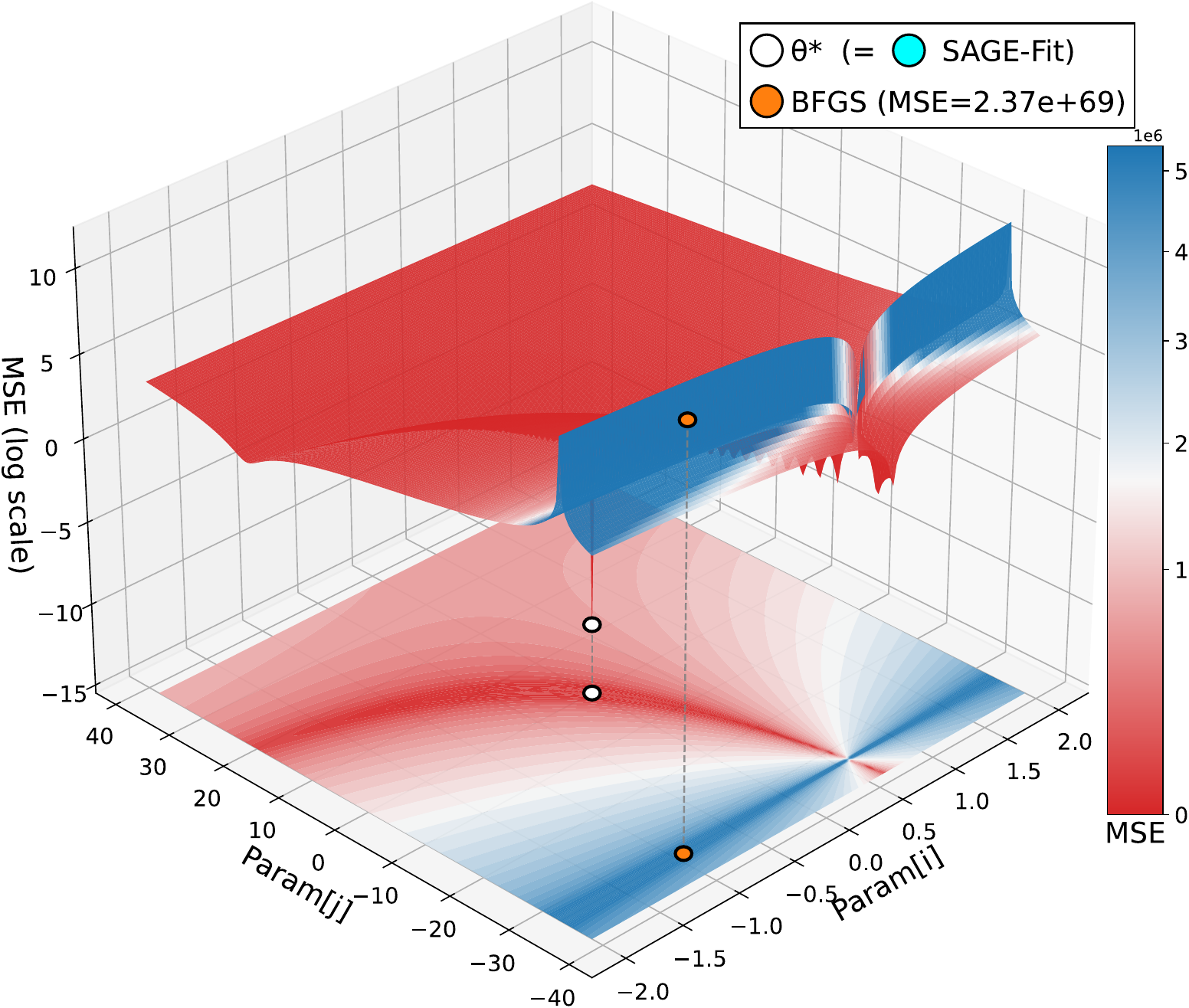} &
            \includegraphics[width=0.484\linewidth]{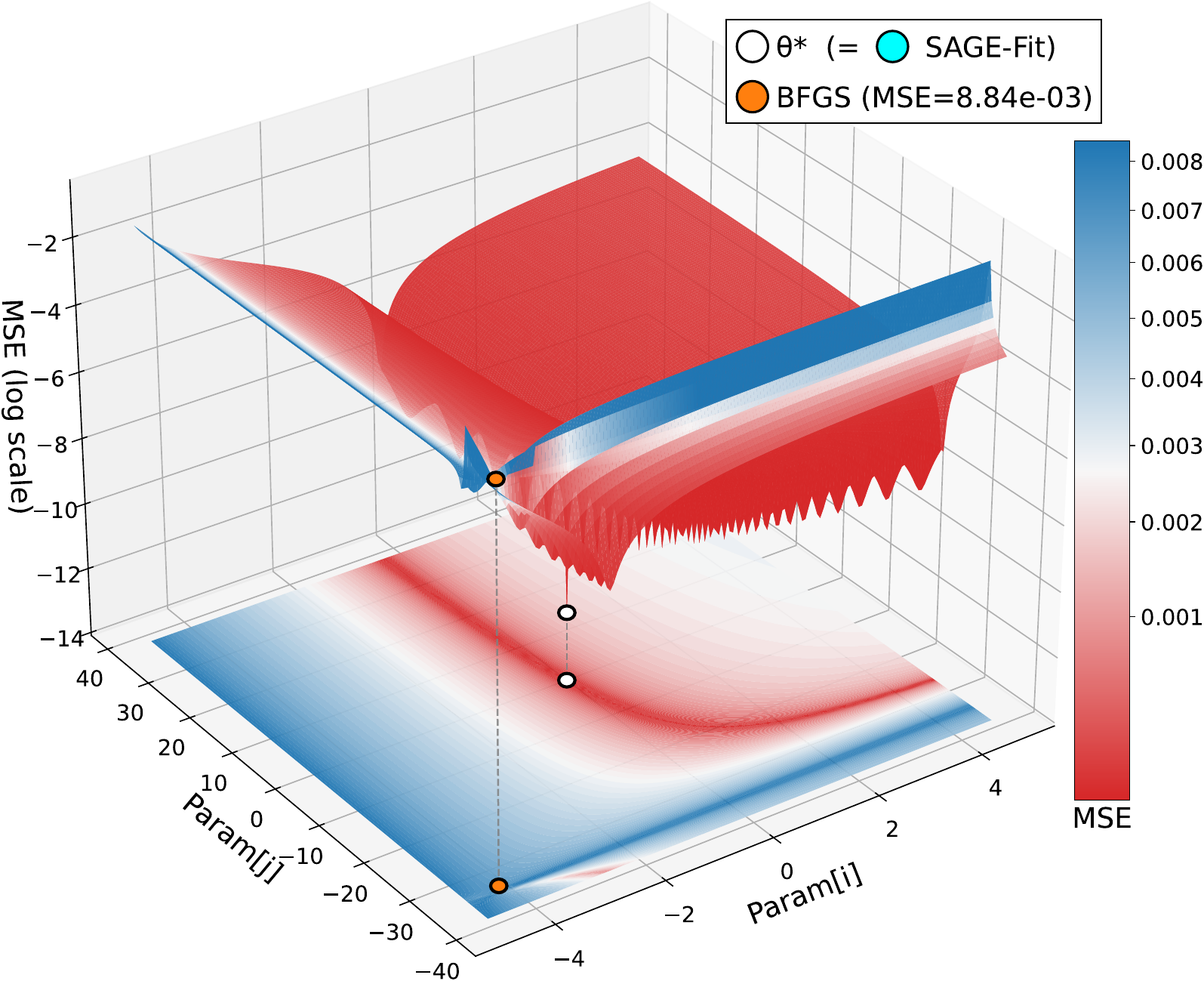}
        \end{tabular}
        \caption{Correct-structure landscapes.}
        \label{fig:landscape_cases_a}
    \end{subfigure}
    \hfill
    \begin{subfigure}[t]{0.49\textwidth}
        \centering
        \setlength{\tabcolsep}{1.5pt}
        \renewcommand{\arraystretch}{0.9}
        \begin{tabular}{@{}cc@{}}
            \includegraphics[width=0.484\linewidth]{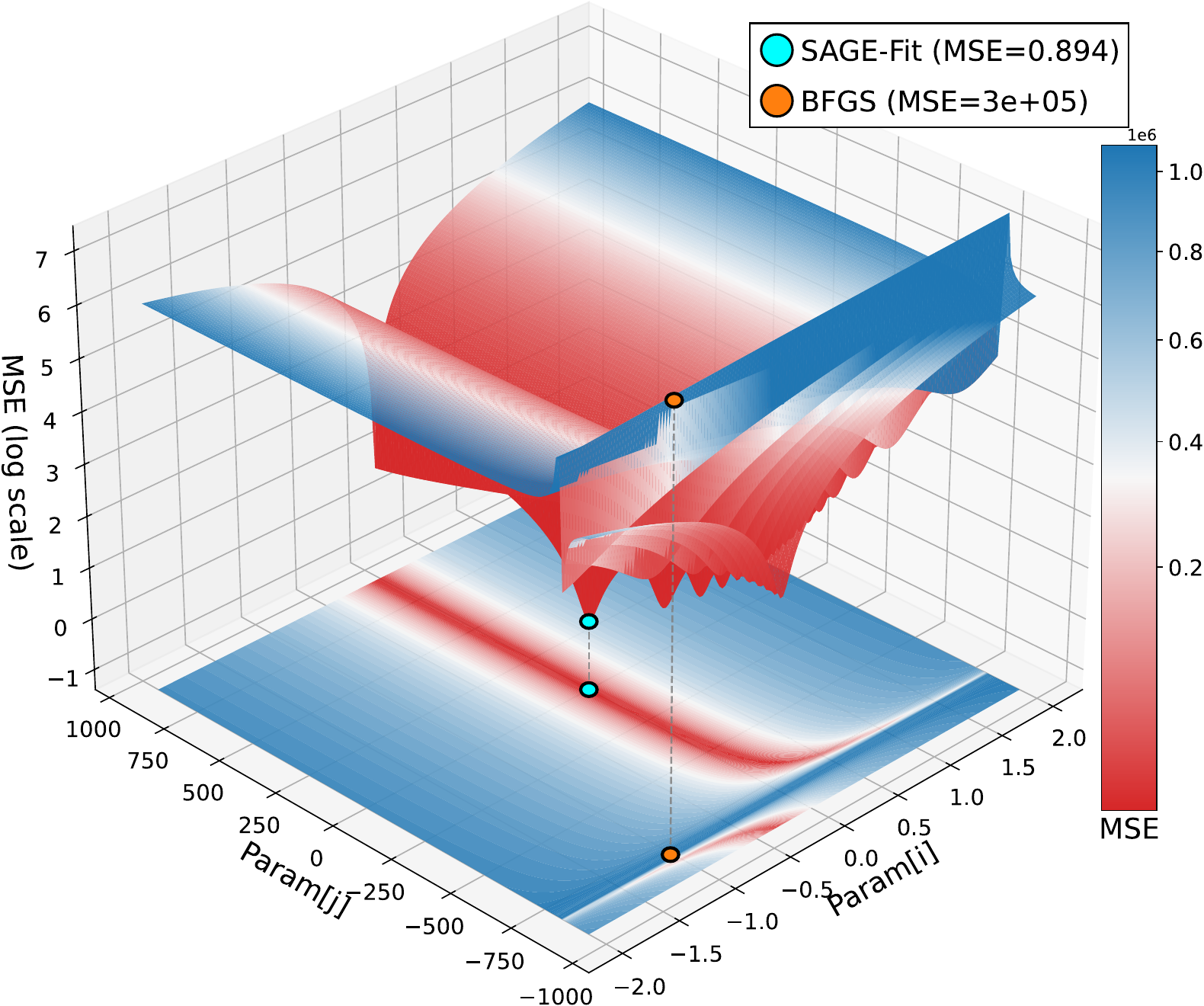} &
            \includegraphics[width=0.484\linewidth]{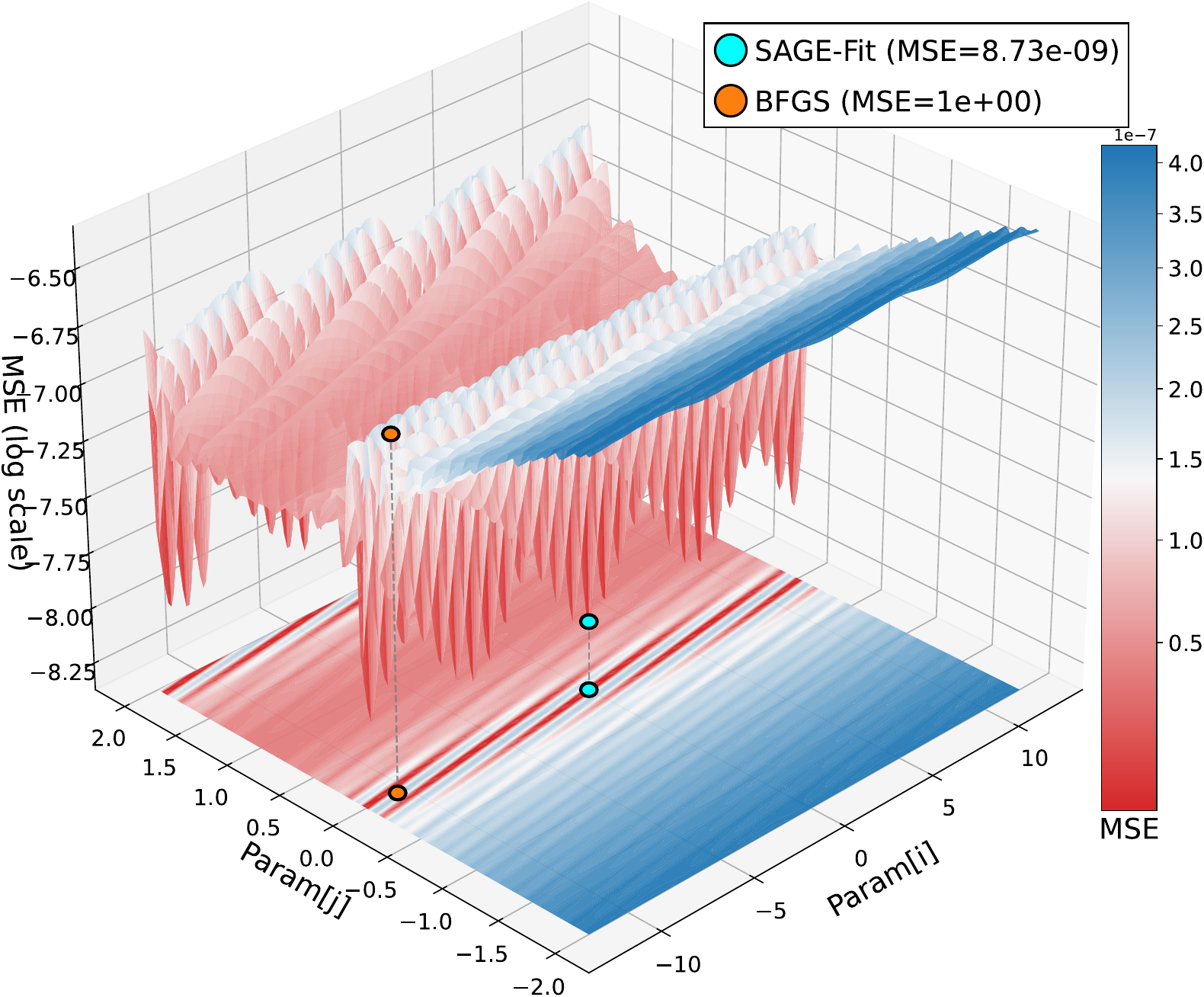}
        \end{tabular}
        \caption{Partially correct structure landscapes.}
        \label{fig:landscape_cases_b}
    \end{subfigure}

\caption{
Representative fitting landscapes in SR.
White, cyan, and orange markers denote ground-truth parameters,
SAGE-Fit solutions, and BFGS solutions, respectively; in \textbf{(A)},
the white and cyan markers coincide.
\textbf{(A)} Correct-structure cases: the optimum is one basin among
multiple local optima, so BFGS may assign a poor score to the correct form.
\textbf{(B)} Partially correct structures: landscapes are more diverse
and unstable, with competing basins and strong initialization sensitivity.
}
    \label{fig:landscape_cases}
\end{figure*}

\section{Anatomy of Fitting Failures}
\label{4}

To understand the root cause of the ``Good Structure, Bad Score'' bottleneck, we must dissect the optimization landscape that triggers these evaluator failures. In SR, fitting failures are predominantly tied to the severe non-convexity induced by nested nonlinear operators. When parameters enter a model linearly, coefficients can be reliably recovered via (regularized) least squares, rendering the fitting outcome largely insensitive to initialization. In contrast, nonlinear parameterization warps the objective into a highly ill-conditioned topography. This renders existing solvers to become initialization-sensitive, leading to premature termination and underestimated scores even for correct structures. To characterize when and why this happens, we analyze the optimization landscapes across two distinct regimes: exact correct structures and partially correct structures.

\subsection{Correct-structure landscapes: SR Scoring Needle-Basin on a Rugged Landscape}
\label{4.1}

We first study the optimization landscape of structures that are structurally identical to the ground-truth equations. Unlike simple linear models, equations with nested nonlinear terms---commonly used in physics to describe oscillatory dynamics or exponential decay---often induce a highly ill-conditioned geometry~\cite{ioannou2026empiricalinvestigationneuralodes,schmidt2009distilling}. We refer to the characteristic geometry arising from the complex coupling of nonlinear parameters as a \emph{Needle-Basin}~\cite{gill1978algorithms}.

\paragraph{The ``Needle-in-a-Basin'' Topology.}
From the perspective of SR scoring, these needle-basins are embedded in a highly rugged loss surface populated with poor stationary points and flat plateaus, as shown in Figure~\ref{fig:landscape_cases_a}. Consequently, single-start local fitters (e.g., BFGS-style methods) are prone to converging to high-error stationary points or stalling on plateaus under conventional initializations~\cite{kommenda2020parameter}. This premature termination triggers stopping criteria early, assigning an unjustifiably poor score to an otherwise correct structure (i.e., an optimizer-induced false negative).

\paragraph{Non-convexity and Multimodality.}
Beyond flat plateaus, the landscape of nonlinear SR structures is inherently multimodal. As illustrated in Figure~\ref{fig:landscape_cases_a}, the parameter space is partitioned into multiple disconnected basins of attraction~\cite{sergeyev2016least,riley2025deflation}. This multimodality often stems from the periodicity of trigonometric operators or the symmetry of exponential terms, creating numerous suboptimal local attractors that compete with the global solution. Standard gradient-based methods lack the mechanism to escape these suboptimal basins once initialized within them.

\paragraph{Stalling and Premature Termination.}
Local fitters commonly used in SR typically rely on gradient-norm--based stopping rules (e.g., $\lVert \nabla L \rVert < \epsilon$). In a rugged needle-basin landscape, once the optimizer traverses into a suboptimal attractor region or a flat plateau, the local gradient magnitudes can diminish rapidly. This satisfies the stopping criterion even though the resulting fitting error remains orders of magnitude higher than the best attainable level for that structure. This phenomenon distorts the score fed back to the outer-loop~\cite{dos2024benchmarking,kommenda2020parameter}, misguiding the structural search process.

\subsection{Landscapes of Partially Correct Structures: Diverse and Unstable Fitting Behaviors}
\label{4.2}
We next investigate the parameter optimization landscapes associated with partially correct structures. Here, a candidate structure is defined as ``partially correct'' if it deviates from the ground-truth equation solely through localized structural modifications (e.g., the omission of a term or the substitution of a local operator). In contrast to the correct structures discussed in Section~\ref{4.1}---which typically exhibit a consistent needle-basin geometry---we observe \textbf{diverse and unstable fitting behaviors} for partially correct structures (see Fig.~\ref{fig:landscape_cases_b}). Specifically, the loss landscapes and resulting optimization trajectories vary markedly, characterized by three recurring phenomena~\cite{kronberger2025effects}. First, low-error regions often manifest as \textbf{needle-basin pockets} or trenches; in certain cross-sections, multiple isolated low-error pockets are observable. Second, we frequently observe \textbf{multiple competitive basins}, where several disconnected basins with comparable minimum errors coexist, implying that the solution yielding a low error is not unique. Finally, the landscape may feature \textbf{wide valleys with high barriers}: although the optimal valley can be relatively broad, it is often separated from suboptimal basins by significant energy barriers or narrow passages.

\section{Method}
\label{sec:method}

The optimization bottlenecks identified in Section~\ref{4} arise because existing optimizers treat equation candidates as black boxes, ignoring their rich structures.
Following the notation in Section~\ref{pre}, let $f$ denote a candidate structure proposed by the outer loop, with parameters $\theta\in\Theta_f\subset\mathbb{R}^{d_f}$. Given the dataset $\mathcal{D}=\{(\bm{x}_i,y_i)\}_{i=1}^{n}$, the ideal fitness score of $f$ is evaluated by the minimum of the empirical fitting loss:
\begin{equation}
\label{eq:ideal-score}
\mathcal{S}^{\star}(f)
=
\min_{\theta\in\Theta_f} L_f(\theta;\mathcal{D})
=
\min_{\theta\in\Theta_f}
\frac{1}{n}\sum_{i=1}^{n}
\left(y_i-f(\bm{x}_i;\theta)\right)^2 .
\end{equation}
However, in practice, the outer loop never observes the true oracle score $\mathcal{S}^{\star}(f)$. Instead, it receives a budget-limited approximation
\begin{equation}
\label{eq:fitted-score}
\widehat{\mathcal{S}}(f)=L_f(\widehat{\theta};\mathcal{D}),
\end{equation}
where $\widehat{\theta}$ denotes the parameters returned by an empirical inner-loop solver under finite computational constraints.
This discrepancy introduces an evaluator gap
\begin{equation}
\label{eq:evaluator-gap}
\Delta(f)=\widehat{\mathcal{S}}(f)-\mathcal{S}^{\star}(f)\ge 0.
\end{equation}
Because existing optimizers struggle to navigate the highly non-convex landscapes, they frequently produce an exceptionally large $\Delta(f)$ even for correct candidates. Consequently, these promising structures are prematurely discarded, resulting in severe optimizer-induced false negatives.

To resolve this, we introduce \textbf{SAGE-Fit} (\textbf{S}tructure-\textbf{A}ware and Semantics-\textbf{G}uided \textbf{E}valuator), an SR-native optimizer designed to explicitly exploit two fundamental priors of symbolic candidates: the \textit{algebraic structural prior} and the \textit{functional semantic prior}. By operationalizing these priors, SAGE-Fit transforms the highly non-convex evaluation process into a structured, three-stage pipeline. First, it utilizes the algebraic prior to analytically decouple conditionally linear parameters, collapsing the search space dimensionality (Module 1, Sec.~\ref{sec:5.2}). Second, it relies on the semantic prior to measure diversity in the function space, enabling robust farthest-point initialization that escapes local minima traps (Module 2, Sec.~\ref{sec:5.3}). Finally, from these high-quality semantic seeds, it applies an SR-specialized Projected Gauss-Newton method for rapid local convergence (Module 3, Sec.~\ref{sec:5.4}). Together, these components yield a unified, high-fidelity evaluator for SR (Sec.~\ref{sec:5.5}).

\subsection{Module 1: Structure-Aware Dimensionality Reduction}
\label{sec:5.2}

\paragraph{The Bottleneck.} 
Traditional optimizers treat all parameters uniformly, entangling conditionally linear coefficients (e.g., the amplitude $a$ in $a\sin(bx+c)$) with non-linear parameters (e.g., the frequency $b$ and phase $c$). This spurious coupling significantly inflates the search dimensionality and exacerbates the ill-conditioning of the loss landscape, creating the deceptive ``needle-in-a-basin'' topography.

\paragraph{Deterministic Dimension Collapse via Variable Projection.}
We exploit the \emph{structural prior} inherent in symbolic expressions: by parsing the abstract syntax tree (AST) of the equation, we automatically identify which parameters are conditionally linear. These are parameters for which the equation becomes affine when all other parameters are held fixed.
Concretely, we traverse the AST of candidate structure $f$ and certify that the parameter $\theta_k$ is conditionally linear if and only if $\frac{\partial f}{\partial\theta_k}$ is independent of $\theta_k$, and 
$f\big|_{\theta_k=0}$ contains no $\theta_k$.
This condition is verified symbolically via an affine-propagation pass over the AST. Parameters satisfying both conditions form the \emph{linear set} $\alpha\in\mathbb{R}^{d_\alpha}$; the remaining parameters constitute the \emph{irreducible non-linear set} $\beta\in\mathbb{R}^{d_\beta}$, with $d_f=d_\alpha+d_\beta$. 
Let $\mathbf{X}=[\bm{x}_1,\ldots,\bm{x}_n]^\top\in\mathbb{R}^{n\times d_x}$, $\mathbf{y}=[y_1,\ldots,y_n]^\top\in\mathbb{R}^{n}$. The candidate structure then admits the separable formulation:
\begin{equation}
\label{eq:separable}
f(\mathbf{X};\alpha,\beta)\;=\;c_f(\mathbf{X};\beta)\;+\;\Phi_f(\mathbf{X};\beta)\,\alpha,
\end{equation}
where \(c_f(\mathbf{X};\beta)\in\mathbb{R}^{n}\) and
\(\Phi_f(\mathbf{X};\beta)\!\in\!\mathbb{R}^{n\times d_\alpha}\) is a basis
matrix whose columns depend on the data and \(\beta\), but remain invariant with
respect to \(\alpha\). Crucially, this decomposition allows the linear basis to
depend on the non-linear parameters. For instance, in \(a\sin(bx+c)\), the
amplitude \(a\) is absorbed into \(\alpha\) and eliminated by a linear solver,
while the frequency and phase \((b,c)\) remain in the projected non-linear
variable set \(\beta\). Appendix~\ref{app:separable_example} provides a concrete
worked example with \(n=100\) data points, explicitly showing the construction
and dimensions of \(c_f(\mathbf{X};\beta)\), \(\Phi_f(\mathbf{X};\beta)\),
\(\alpha\), and \(\beta\).

For any fixed $\beta$, the optimal $\alpha$ is obtained in closed form via linear least squares:
\begin{equation}
\label{eq:alpha-star}
\resizebox{\columnwidth}{!}{$\displaystyle
\alpha^\star(\beta)\;=\;\big(\Phi_f(\mathbf{X};\beta)^{\!\top}\Phi_f(\mathbf{X};\beta)\big)^{\!\dagger}\Phi_f(\mathbf{X};\beta)^{\!\top}\big(\mathbf{y}-c_f(\mathbf{X};\beta)\big),
$}
\end{equation}
where $(\cdot)^\dagger$ denotes a rank-revealing pseudo-inverse with Tikhonov regularization applied when $\Phi_f^{\!\top}\Phi_f$ is rank-deficient. This step involves a brief, deterministic linear solution process, which is never delegated to a non-linear optimizer. The original $d_f$-dimensional problem collapses to the projected non-linear program
\begin{equation}
\label{eq:proj-loss}
\resizebox{\columnwidth}{!}{$\displaystyle
\min_{\beta\in\mathcal{B}_f}\ L_{\mathrm{proj}}(\beta)
\;=\;\frac{1}{n}\big\|\mathbf{y}-c_f(\mathbf{X};\beta)-\Phi_f(\mathbf{X};\beta)\,\alpha^\star(\beta)\big\|_2^2,
$}
\end{equation}
whose argument resides on an irreducible non-linear manifold of dimension $d_\beta\!<\!d_f$.

\subsection{Module 2: Semantics-Guided Initialization via Function-Space Farthest-Point Sampling}
\label{sec:5.3}

\paragraph{The Bottleneck.} As characterized in Section~\ref{4}, the optimization landscape of SR is usually multimodal, featuring disconnected basins of attraction and local optimum traps. To escape local minima, multi-start strategies are essential. However, traditional approaches commonly sample starting points uniformly within the \emph{parameter space}. This strategy is fundamentally problematic for SR: the distance in the parameter space is highly deceptive. For example, $\sin(x)$ and $\sin(x+2\pi)$ are semantically identical despite possessing vastly different parameter coordinates. Random sampling in the parameter space thus wastes computational budget on redundant starting points that collapse into the same basin, a phenomenon we term \emph{basin collapse}.

\paragraph{Diversity in Function Space, Not Parameter Space.}
To prevent basin collapse, SAGE-Fit evaluates diversity based on the input-output mappings of candidates. Specifically, we introduce \emph{Function-Space Farthest-Point Sampling} (FS-FPS), which ensures that starting points are maximally diverse in terms of their \emph{observed behavior} on the training data, rather than their parameter coordinates. 
The algorithm proceeds in three steps:
\begin{enumerate} \item \textbf{Oversampling}: Generate $M$ candidate parameter vectors by sampling around the warm initialization within the projected $\beta$ space (following the VarPro dimension reduction). \item \textbf{Function-space evaluation}: For each candidate $\beta_m$, compute the predicted output $\mathbf{f}_m\!=\!f(\mathbf{X};\alpha^\star(\beta_m),\beta_m)$ on the training set. This procedure maps each parameter vector to its \emph{semantic fingerprint}, which is the actual function it represents. \item \textbf{Farthest-point selection}: Within the space of output vectors $\{\mathbf{f}_m\}$, greedily select $K$ points that are maximally separated under the normalized Euclidean distance \begin{equation} \label{eq:fs-distance} d_{\mathrm{sem}}(\mathbf{f}_i,\mathbf{f}_j)\;=\;\frac{\|\mathbf{f}_i-\mathbf{f}_j\|_2}{\|\mathbf{y}\|_2+\epsilon}. \end{equation} The first point is the candidate with the lowest training loss; subsequent points are chosen to maximize the minimum distance to all previously selected points. \end{enumerate} The concrete values of $M$ and $K$ used in our experiments are reported in Appendix~\ref{app:hyperparameters}. This procedure guarantees that the $K$ selected starting points exhibit maximal \emph{behavioral diversity}: they produce genuinely different predictions on the training data. Consequently, they cover distinct regions of the function space and significantly reduce the risk of basin collapse.

\subsection{Module 3: SR-Specialized Local Attraction via Projected Gauss--Newton}
\label{sec:5.4}

\paragraph{The Bottleneck.} Even after dimension reduction and semantic initialization, the projected non-linear manifold $L_{\mathrm{proj}}(\beta)$ remains highly non-convex due to trigonometric periodicity, exponential plateaus, and nested non-linearities. Standard gradient descent methods are prohibitively slow on such optimization landscapes.

\paragraph{Curvature-Aware Descent on the Projected Manifold.}
We employ a \emph{Projected Gauss--Newton} step with Trust Region Reflective (TRF) damping, specifically tailored for the SR setting.
Define the projected residual $\tilde r(\beta)=\mathbf{y}-c_f(\mathbf{X};\beta)-\Phi_f(\mathbf{X};\beta)\,\alpha^\star(\beta)$ and the corresponding projected Jacobian $\tilde J(\beta)=\partial\tilde r/\partial\beta\!\in\!\mathbb{R}^{n\times d_\beta}$. Following the Golub--Pereyra construction~\cite{golub2003separable}, the parameter vector $\alpha^\star(\beta)$ is re-solved at every evaluation of $\beta$ via Eq.~\eqref{eq:alpha-star}. Consequently, $\tilde r$ is a function of solely $\beta$, and $\tilde J$ implicitly accounts for the dependence of $\alpha^\star$ on $\beta$. In this implementation, $\tilde J$ is obtained through a central finite-difference pass over $\beta$, with $\alpha^\star$ recomputed by the same rank-guarded least squares solver at each finite-difference probe. The projected Levenberg--Marquardt step is formulated as follows:
\begin{equation}
\label{eq:proj-gn}
\Delta\beta\;=\;-\Big(\tilde J^{\!\top}\tilde J\;+\;\lambda_\beta\big(\operatorname{diag}(\tilde J^{\!\top}\tilde J)+\eta I\big)\Big)^{-1}\tilde J^{\!\top}\tilde r,
\end{equation}
with a damping parameter $\lambda_\beta\!\ge\!0$ and a small numerical floor $\eta\!>\!0$ on the diagonal, which prevents zero values in cases where an entry in $\beta$ exerts no leverage on $\tilde r$.

Two properties make this step particularly well suited to SR optimization landscapes. First, the Hessian-like matrix is of dimension $d_\beta\!\times\!d_\beta$ rather than $d_f\!\times\!d_f$: the curvature observed by the solver represents the curvature of the irreducible manifold alone, free from the spurious coupling that conditionally linear coefficients would otherwise inject into a full Newton system. Second, the scale anisotropy across the entries of $\beta$ (e.g.,\ $\sin(\omega x)$ versus\ $x^k$) is normalized in a coordinate-wise manner by the term $\operatorname{diag}(\tilde J^{\!\top}\tilde J)$.

\subsection{The SAGE-Fit Evaluator}
\label{sec:5.5}

The three aforementioned modules are integrated into a unified evaluator (Algorithm~\ref{alg:sagefit}) that takes a candidate structure $f$ as input and returns a faithful score $\widehat{\mathcal{S}}(f)$, along with the fitted parameters $(\hat\alpha,\hat\beta)$.

\paragraph{Discussion.}
SAGE-Fit renders the inner loop \emph{prior-aware} along two orthogonal axes. The \emph{structural prior} (Module 1) exploits the algebraic structure of the AST to deterministically collapse the search space from $d_f$ dimensions to $d_\beta\!<\!d_f$ dimensions, thereby eliminating spurious linear-to-non-linear coupling. The \emph{semantic prior} (Module 2) ensures that the diversity of multiple starting points is measured in the function space rather than the parameter space. This approach prevents basin collapse and guarantees that each starting point explores a genuinely distinct region of the optimization landscape. Subsequently, Module 3 provides rapid, curvature-aware local convergence originating from these high-quality seed points.

Consequently, this three-module pipeline directly addresses the two failure regimes characterized in Section~\ref{4}. Specifically, the needle-in-a-basin topographies driven by linear-to-non-linear coupling are alleviated by the VarPro projection (Module 1), whereas the residual multimodality on the projected manifold is mitigated through semantic diversity during initialization (Module 2), followed by rapid local attraction (Module 3). In contrast to traditional optimizers that treat all parameters and starting points uniformly, SAGE-Fit leverages the unique structure inherent in SR to achieve both high reliability and computational efficiency.

\section{Experiment}

\begin{table*}[t]
\centering
\normalsize
\setlength{\tabcolsep}{4.0pt}
\renewcommand{\arraystretch}{1.25}

\resizebox{\textwidth}{!}{%
\begin{tabular}{l
>{\columncolor{gray!15}}c cc
>{\columncolor{gray!15}}c cc
>{\columncolor{gray!15}}c cc
>{\columncolor{gray!15}}c cc}
\toprule
\textbf{Models}
& \multicolumn{12}{c}{\textbf{LSR-Synth}} \\
\cmidrule(lr){2-13}
& \multicolumn{3}{c}{Chemistry}
& \multicolumn{3}{c}{Biology}
& \multicolumn{3}{c}{Physics}
& \multicolumn{3}{c}{Material Science} \\
\cmidrule(lr){2-4}\cmidrule(lr){5-7}\cmidrule(lr){8-10}\cmidrule(lr){11-13}
& SA(\%)$\uparrow$ & Acc$_{0.1}$(\%)$\uparrow$ & NMSE$\downarrow$
& SA(\%)$\uparrow$ & Acc$_{0.1}$(\%)$\uparrow$ & NMSE$\downarrow$
& SA(\%)$\uparrow$ & Acc$_{0.1}$(\%)$\uparrow$ & NMSE$\downarrow$
& SA(\%)$\uparrow$ & Acc$_{0.1}$(\%)$\uparrow$ & NMSE$\downarrow$ \\
\midrule

PySR
& 2.78 & 37.14 & 5.30e-4
& 13.04 & 19.05 & 3.67e-3
& 0.00 & 16.00 & 4.80e-3
& 4.00 & 11.76 & 1.90e-5 \\
PySR+
& \textbf{8.33} & \textbf{43.33} & \textbf{3.88e-4}
& \textbf{29.17} & \textbf{45.83} & \textbf{3.51e-3}
& \textbf{2.27} & \textbf{52.63} & \textbf{2.92e-3}
& \textbf{16.00} & \textbf{41.67} & \textbf{1.11e-5} \\
\midrule

uDSR
& 11.11 & 54.29 & 4.05e-7
& 0.00 & 20.83 & 7.76e-2
& 0.00 & 7.50 & 1.19e-2
& 0.00 & 13.04 & 1.20e-3 \\
uDSR+
& \textbf{13.89} & \textbf{64.89} & \textbf{1.22e-7}
& \textbf{4.17} & \textbf{23.81} & \textbf{2.37e-2}
& \textbf{2.27} & \textbf{12.50} & \textbf{4.82e-3}
& 0.00 & \textbf{15.79} & \textbf{1.06e-3} \\
\midrule

LaSR
& 2.78 & 25.00 & 1.50e{-3}
& 0.00 & 0.00 & 3.47e{-1}
& 2.27 & 6.82 & 5.95e{-3}
& 0.00 & 4.00 & 1.50e{-3} \\
LaSR+
& \textbf{5.56} & \textbf{27.78} & \textbf{1.38e{-3}}
& \textbf{16.67} & \textbf{8.33} & \textbf{1.02e{-3}}
& \textbf{6.81} & \textbf{9.09} & \textbf{3.85e{-3}}
& \textbf{4.00} & \textbf{8.00} & \textbf{6.02e{-4}} \\
\midrule

LLM-SR
& 8.33 & 33.33 & 1.91e{-5}
& 16.67 & 41.67 & 4.54e{-5}
& 4.55 & 36.84 & 1.04e{-4}
& 4.00 & 11.36 & 2.31e{-5} \\
LLM-SR+
& \textbf{11.11} & \textbf{60.61} & \textbf{1.50e{-5}}
& \textbf{20.80} & 41.67 & \textbf{1.84e{-5}}
& \textbf{11.36} & \textbf{48.00} & \textbf{9.23e{-5}}
& \textbf{8.00} & \textbf{15.91} & \textbf{2.05e{-5}} \\

\bottomrule
\end{tabular}%
}
\caption{Overall performance on LLM-SRBench.}
\label{tab:tb2}
\end{table*}

\subsection{Experimental Setup}

\paragraph{Benchmark Datasets.}
To rigorously evaluate the performance of SAGE-Fit under conditions that mimic real-world scientific discovery, we utilize the LSR-Synth dataset from the LLM-SRBench suite~\cite{shojaee2025llmsrbenchnewbenchmarkscientific}, which comprises 129 distinct equation discovery problems. This dataset provides two critical conditions for our study. First, it mitigates the risk of data contamination and rote memorization often observed in LLMs by constructing equations through the composition of known scientific terms with synthetic, yet plausible, novel terms. Second, the dataset reflects the structural complexity of real-world scientific models. The constituent equations feature diverse combinations of non-linear operators that naturally yield the highly non-convex optimization landscapes. This inherent complexity provides a rigorous and realistic environment to evaluate the robustness of inner-loop parameter solvers.
More details about the dataset construction are provided in Appendix~\ref{data}.

\paragraph{Baselines.}
We evaluate SAGE-Fit as a plug-and-play evaluator by integrating it into four representative SR frameworks: \textbf{PySR}, an evolutionary SR system with symbolic simplification~\cite{cranmer2023interpretablemachinelearningscience}; \textbf{uDSR}, which uses a pre-trained Transformer and neural-guided decoding to generate expression trees~\cite{landajuela2022unified}; \textbf{LaSR}, which combines LLM generation with evolutionary search through a learned symbolic concept library~\cite{grayeli2024symbolic}; and \textbf{LLM-SR}, which searches over equation skeletons represented as Python programs~\cite{shojaee2025llmsr}. Each baseline is compared with a corresponding ``+'' variant (e.g., PySR+), in which only the original parameter optimizer is replaced by SAGE-Fit, while the outer-loop search remains unchanged. All methods are evaluated under a time limit of 1800 seconds per problem. Each baseline--``+'' pair uses the same random seed, outer-loop hyperparameters, search configuration, and hardware environment. For LaSR and LLM-SR, we use Qwen2.5-7B as the backbone and keep the model deployment, decoding parameters, prompts, and search settings identical across paired runs. Detailed implementation settings are provided in Appendix~\ref{data}.

\paragraph{Evaluation Metrics.}
We evaluate all methods using three standard metrics: numerical fit, prediction accuracy, and symbolic correctness. We report the Normalized Mean Squared Error (NMSE) to assess the overall numerical accuracy, where $\bar{y}$ denotes the mean of the target values. We further report the Accuracy to Tolerance ($\mathrm{Acc}_{\tau}$), which measures the proportion of predictions whose relative errors are within a tolerance $\tau$. The two metrics are defined as
\[
\resizebox{\columnwidth}{!}{$
\displaystyle
\mathrm{NMSE}
=
\frac{\sum_{i=1}^{N}(\hat{y}_i-y_i)^2}
     {\sum_{i=1}^{N}(y_i-\bar{y})^2},
\qquad
\mathrm{Acc}_{\tau}
=
\frac{1}{N}
\sum_{i=1}^{N}
\mathbb{I}\!\left(
\left|
\frac{\hat{y}_i-y_i}{y_i}
\right|
\le \tau
\right)
$}
\]
Finally, to evaluate the structural quality of the discovered expressions, we adopt the Symbolic Accuracy (SA) metric. SA measures whether the discovered equation recovers the correct symbolic form, specifically checking whether the found equation is mathematically equivalent to the ground-truth equation up to fitted constants.
Meanwhile, SA is a binary metric for symbolic recovery, assigning zero credit to any structural deviation from the ground truth. Consequently, SA scores are typically low, and even modest gains indicate meaningful improvements in exact recovery.

\subsection{Main Results}

\paragraph{Performance Comparison.}
Table~\ref{tab:tb2} summarizes the main results. For a fair comparison, we apply SAGE-Fit to the final equations produced by all baselines, including PySR, uDSR, LaSR, and LLM-SR, and compute the baseline rows after this post-hoc refinement. The refinement consistently reduces NMSE, confirming the convergence capability of SAGE-Fit. Standardizing final parameter fitting separates gains from improved final tuning and improved search guidance, so differences between each baseline and its ``+'' variant reflect the ability of SAGE-Fit to guide the outer loop. Under this protocol, all ``+'' variants improve structure discovery. PySR+ raises SA from 2.78\% to 8.33\% ($3\times$), while uDSR+ achieves the highest Chemistry SA of 13.89\%. These gains show that more faithful inner-loop optimization mitigates the Good Structure, Bad Score problem by retaining correct structures that suboptimal fitting may discard.

The ``+'' variants also outperform the post-hoc-refined baselines in both $\mathrm{Acc}_{0.1}$ and NMSE, indicating that SAGE-Fit guides the search toward more accurate and generalizable structures instead of merely improving the parameters of inferior ones. In particular, LLM-SR+ nearly doubles accuracy from 33.33\% to 60.61\% in the first group, demonstrating the benefit of combining LLM-generated hypotheses with precise evaluation. The consistent gains across metrics and baseline methods under the same time budget establish SAGE-Fit as an effective plug-and-play evaluator that improves existing SR frameworks without modifying their outer-loop search logic.

\paragraph{Trajectory-Level Analysis.}
A fine-grained analysis of the LLM-SR trajectories further shows that,
across all 129 tasks, SAGE-Fit reduces the final NMSE by an average of
$3.22$ orders of magnitude. On average, $22.16\%$ of the discarded
candidate equations would outperform their corresponding incumbents
after refitting, with this rate reaching $65.75\%$ in the most affected
trajectory. This directly supports the ``Good Structure, Bad Score''
hypothesis, showing that inaccurate parameter fitting can eliminate
promising expressions before they influence subsequent search.
Detailed definitions and results are provided in
Appendix~\ref{app:trajectory_diagnostics}.

\subsection{Optimizer Comparison and Ablation on Progress-Associated Candidates}
\label{sec:optimizer_ablation}
\label{sec:spb}

The preceding trajectory-level analysis shows that inaccurate inner-loop evaluation can distort closed-loop search. To isolate this effect at the candidate level, we construct the Structure-Progress Bank (SPB), a diagnostic benchmark that, unlike standard SR benchmarks focused on final outputs, contains candidate equations from real search trajectories with observable score improvements. SPB includes 200 candidates from 20 problems, with 10 progress-associated candidates per problem. Each candidate corresponds to a score-improvement event, forming a targeted set of structures already identified as potentially useful by the outer loop. Each is represented as a callable expression with free continuous parameters and contains 2 to 11 parameters, with a median of 7. All evaluators use the same cold-start initialization $\theta_0$, allowing SPB to assess recovery of useful fits without trajectory-specific warm starts.

We compare four evaluators: single-start L-BFGS-B ($\text{BFGS\_single}$), a common SR fitting routine; multi-start L-BFGS-B ($\text{BFGS\_multi}$); SAGE-Fit w/o FS-FPS, an ablated variant that replaces function-space farthest-point start selection with random multi-start initialization; and the full SAGE-Fit.

\paragraph{Empirical reference score and SPB lost rate.}
For each candidate $f_i$, we define the empirical reference score as
$\widehat{\mathcal{S}}_{\mathrm{ref}}(f_i)
=\min_{e\in\mathcal{E}}\widehat{\mathcal{S}}_{e}(f_i)$,
where $\mathcal{E}$ comprises BFGS\_single, BFGS\_multi,
SAGE-Fit w/o FS-FPS, and SAGE-Fit. This score is the lowest
NMSE attained within the comparison set and is used to quantify
relative fitting failures among the evaluated methods.

An evaluator is considered to miss a candidate if its fitted NMSE is substantially worse than this empirical reference:
\begin{equation*}
\resizebox{\columnwidth}{!}{$
\displaystyle
\mathrm{LostRate}(e;\tau)
=
\#\!\left\{
f_i\in {\mathrm{SPB}}:
\widehat{\mathcal{S}}_{e}(f_i)
>
\tau\,\widehat{\mathcal{S}}_{\mathrm{ref}}(f_i)
\right\}
\big/
|{\mathrm{SPB}}|
$}
\end{equation*}
Here, $\tau$ is a fixed threshold multiplier used to identify substantial
fitting failures. A high lost rate indicates that an evaluator frequently
assigns poor scores to candidates that other evaluators fit substantially
better. The SPB diagnostic hyperparameters are reported in
Appendix~\ref{app:hyperparameters}.

 \begin{table}[t]
      \centering
      \resizebox{\columnwidth}{!}{
      \begin{tabular}{lcccc}
          \toprule
          \textbf{Evaluator}
          & \textbf{Lost Rate} $\downarrow$
          & \multicolumn{2}{c}{\textbf{log NMSE} $\downarrow$}
          & \textbf{Time} $\downarrow$ \\
          \cmidrule(lr){3-4}
          &
          & \textbf{Mean}
          & \textbf{Median}
          & \\
          \midrule
          SAGE-Fit
          & \textbf{9.5\%}
          & \textbf{-3.863}
          & \textbf{-3.805}
          & 0.57s \\
          SAGE-Fit w/o FS-FPS
          & 15.0\%
          & -3.786
          & -3.705
          & 0.96s \\
          \text{BFGS\_multi}
          & 54.5\%
          & -2.470
          & -2.508
          & 0.37s \\
          \text{BFGS\_single}
          & 70.0\%
          & -1.460
          & -1.005
          & \textbf{0.16s} \\
          \bottomrule
      \end{tabular}
      }

      \caption{
      Candidate-level optimizer comparison and ablation on the Structure-Progress Bank.
      Time denotes the median fitting time per candidate.
      \label{tab:spb_main}
      }
  \end{table}

Table~\ref{tab:spb_main} first evaluates whether standard BFGS-based routines can resolve the observed fitting failures. The single-start setting represents the common low-cost configuration in SR pipelines, whereas the multi-start setting allocates additional restarts to the same optimizer. Although \text{BFGS\_multi} improves over \text{BFGS\_single}, it remains substantially worse than SAGE-Fit in both lost rate and fitted error. This gap shows that evaluator failure cannot be explained by restart count alone. \text{BFGS\_multi} still operates in the original parameter space, where conditionally linear and nonlinear parameters remain coupled and inflate the effective search dimension, causing random starts to revisit difficult or redundant regions. SAGE-Fit instead removes conditionally linear parameters through structure-aware projection and selects diverse function-space starts through semantics-guided initialization. Its pairwise advantage over \text{BFGS\_multi} is broad rather than driven by a few outliers, indicating consistently higher candidate-level fitting fidelity.

The comparison with SAGE-Fit w/o FS-FPS isolates the role of semantics-guided initialization in the fidelity--cost trade-off. This variant uses the same projected solver and number of final nonlinear solver calls as SAGE-Fit, but replaces FS-FPS with random multi-start selection. The full method achieves a lower lost rate, a slightly better mean log NMSE, and a lower observed median fitting time, showing that FS-FPS selects higher-quality starts and avoids redundant refinements. Although SAGE-Fit costs more than the BFGS-based evaluators, it substantially reduces missed candidates and fitted errors, yielding more reliable feedback per fitting budget. Representative rescue cases are provided in Appendix~\ref{app:spb_rescue_cases}. Overall, SPB shows that unreliable fitting can suppress candidates already identified by the outer loop as useful, beyond the effects of restart count or solver choice. By combining structure-aware projection with semantics-guided initialization, SAGE-Fit provides more faithful candidate scores under a competitive fitting cost. This candidate-level evidence complements the earlier trajectory-level lost-rate analysis and supports the need for high-fidelity evaluation to preserve promising structures during closed-loop SR search.

\section{Conclusion}

This work addresses a critical bottleneck in SR: the Good Structure, Bad Score phenomenon. Analysis of the optimization landscape reveals that the intrinsic non-convexity and needle-in-a-basin topology of physical equations frequently cause standard continuous optimizers to fail. These failures underestimate achievable scores and lead search algorithms to discard promising structures. To address this issue, we propose SAGE-Fit, an SR-native parameter optimizer that explicitly exploits the structural and semantic priors of symbolic expressions. By combining structure-aware dimensionality reduction, semantics-guided initialization, and a specialized Projected Gauss--Newton method, SAGE-Fit overcomes the limitations of uniform parameter treatment. As a plug-and-play module, it delivers superior performance across existing SR frameworks. Extensive experiments show that upgrading the inner-loop optimizer consistently improves both symbolic accuracy and numerical fidelity without modifying outer-loop search logic. Refit-based lost-rate evaluations further reveal that optimization failures alone discard up to $65\%$ of structurally superior candidates in complex domains, highlighting the prevalence of optimizer-induced false negatives. By shifting attention from hypothesis generation to accurate evaluation, this study establishes robust parameter optimization as a practical mechanism for improving SR reliability and encourages the community to re-examine this previously underemphasized bottleneck in modern SR research.

\bibliography{ref}

\clearpage
\onecolumn
\appendix

\section{Related Work}

Given the prevailing adoption of the bi-level optimization framework in modern SR, we structure our review of prior literature through the lens of this decomposition. Specifically, we categorize existing methodologies into two complementary tracks: (i) outer-loop structure search algorithms, and (ii) inner-loop parameter optimization strategies.

\subsection{Structure Search}

Early SR methods primarily relied on deterministic modeling and sparse regression, such as FFX~\cite{mcconaghy2011ffx, vaddireddy2019equation, kammerer2022symbolic} and SINDy~\cite{brunton2016discovering, brunton2016sparse,udrescu2020ai}. These methods exhibit high efficiency and stability when the hypothesis space is constrained to a predefined library of basis functions; however, their expressive power is largely limited by the design of the function library.

To expand the searchable function space, genetic programming (GP)-based SR methods have been extensively studied~\cite{koza1994genetic, schmidt2009distilling, virgolin2019linear}. These methods typically represent expressions as trees and perform global search over the symbolic composition space via evolutionary operators such as mutation and crossover. Building upon this paradigm, methods such as PySR~\cite{cranmer2023interpretablemachinelearningscience} further adopt multi-objective formulations for candidate selection, leading to improved search efficiency and quality.

With the advancement of neural network methodologies, SR has increasingly been combined with neuro-symbolic and search-driven approaches. For instance, Equation Learner (EQL) replaces standard neural network activation functions with interpretable mathematical operators~\cite{martius2016extrapolation,sahoo2018learning,dong2024evolving}, enabling the learning of symbolically meaningful functional connections. Other methods, including DSR~\cite{petersen2021deep}, DSO~\cite{landajuela2022unified, hayes2025deep}, and DSN~\cite{pmlr-v235-li24ap}, formulate SR as a sequential decision-making problem and employ reinforcement learning to explore the expression space. In addition, several studies integrate Monte Carlo Tree Search (MCTS) with deep generative models to enhance global exploration and the stability of symbolic expression search~\cite{kamienny2023deep,huang2025improving}.

More recently, SR has witnessed growing interest in large language model (LLM)-driven approaches~\cite{shojaee2025llmsr,merler-etal-2024-context,hua2025finetuninglargelanguagemodel,grayeli2024symbolic}. Benefiting from strong capabilities in code generation and symbolic reasoning, these models are used to propose symbolic expressions or programmatic equation templates. Furthermore, retrieval-augmented methods such as RAG-SR incorporate external or historical expressions through retrieval-augmented generation mechanisms, thereby improving the stability and quality of generated symbolic models~\cite{zhang2025ragsr, srllm}.

\subsection{Parameter Optimization}

Early GP-based SR often performed constant fitting using local or stochastic search methods such as hill climbing~\cite{koza1994genetic} and simulated annealing~\cite{stinstra2008metamodeling, kantor2021simulated}. To improve robustness under highly non-convex objectives, subsequent work introduced population-based, gradient-free optimizers, including genetic algorithms~\cite{alonso2009evolution}, particle swarm optimization~\cite{loebl2018continuous, sheta2023evolutionary}, and differential evolution~\cite{cerny2008using}. These derivative-free approaches are generally less sensitive to poor local geometry, but can be sample-inefficient or yield suboptimal numerical accuracy.

With the increasing prevalence of bi-level SR pipelines, the inner-loop fitting problem is typically solved by continuous optimizers specialized for regression-style objectives. Many systems rely on fast local optimizers such as quasi-Newton methods (e.g., BFGS) or simplex-based methods (e.g., Nelder--Mead) for rapid error reduction~\cite{cranmer2023interpretablemachinelearningscience,petersen2021deep}. When the objective can be cast as (or approximated by) nonlinear least squares, least-squares solvers such as Levenberg--Marquardt are also commonly used for refinement~\cite{burlacu2020operon}. In neural-network-based SR, parameter adjustment is typically carried out during training via gradient-based optimization, while additional local refinement may be applied to obtain precise constants for the final symbolic form~\cite{martius2016extrapolation,sahoo2018learning,pmlr-v235-li24ap}.

These choices are reflected across representative SR systems: 
PySR fits constants using BFGS or Nelder--Mead~\cite{cranmer2023interpretablemachinelearningscience}; 
Operon uses Levenberg--Marquardt least-squares refinement~\cite{burlacu2020operon}; 
DSR refits constants with BFGS and uses the optimized error as the reward signal for structure search~\cite{petersen2021deep}; 
and LLM-SR updates free parameters through automatic differentiation with optimizers such as Adam or BFGS~\cite{shojaee2025llmsr}. 
These methods show that parameter fitting is already a standard component of modern SR pipelines, but it is usually implemented as a standard numerical subroutine on the original parameterization, with limited use of equation-level priors.

\section{Algorithmic Pseudocode for SAGE-Fit Evaluator}

\begin{algorithm}[H]
\caption{SAGE-Fit Evaluator}
\label{alg:sagefit}
\begin{algorithmic}[1]
\Require Candidate structure $f$, data $(\mathbf{X},\mathbf{y})$, bounds $[\ell,u]$, oversampling parameter $M\!=\!100$, number of starting points $K\!=\!8$
\Ensure Faithful score $\widehat{\mathcal{S}}(f)$ and parameters $(\hat\alpha,\hat\beta)$
\State \textbf{Module 1: Structure-Aware Dimensionality Reduction}
\State $(\alpha,\beta,\Phi_f,c_f)\gets\Call{ASTAffinePropagation}{f}$ \Comment{Eq.~\eqref{eq:separable}}
\State \textbf{Module 2: Semantics-Guided Initialization}
\State Sample $M$ candidates $\{\beta_m\}_{m=1}^{M}$ around the warm-start initialization within $[\ell,u]$
\For{$m=1,\dots,M$}
  \State $\mathbf{f}_m\gets f(\mathbf{X};\alpha^\star(\beta_m),\beta_m)$ \Comment{Function-space evaluation}
\EndFor
\State $\{\beta_{i_1},\dots,\beta_{i_K}\}\gets\Call{FarthestPointSampling}{\{\mathbf{f}_m\},K}$ \Comment{Eq.~\eqref{eq:fs-distance}}
\State \textbf{Module 3: SR-Specialized Local Attraction}
\For{$k=1,\dots,K$ \textbf{in parallel}}
  \State $\hat\beta_k\gets\Call{ProjectedTRF}{\beta_{i_k},L_{\mathrm{proj}}}$ \Comment{Eqs.~\eqref{eq:alpha-star},\eqref{eq:proj-loss},\eqref{eq:proj-gn}}
\EndFor
\State $\hat\beta\gets\arg\min_{\beta\in\{\hat\beta_1,\dots,\hat\beta_K\}}L_{\mathrm{proj}}(\beta)$
\State $\hat\alpha\gets\alpha^\star(\hat\beta)$ via Eq.~\eqref{eq:alpha-star}
\State $\widehat{\mathcal{S}}(f)\gets L_{\mathrm{proj}}(\hat\beta)$
\State \Return $(\widehat{\mathcal{S}}(f),\hat\alpha,\hat\beta)$
\end{algorithmic}
\end{algorithm}

\section{Hyperparameter Values} \label{app:hyperparameters} Table~\ref{tab:hyperparameters} summarizes the fixed hyperparameter values used in SAGE-Fit and the SPB diagnostic analysis. These values are kept unchanged across the corresponding paired comparisons. \begin{table}[H] \centering \caption{ Hyperparameter values used in SAGE-Fit and SPB diagnostics. } \label{tab:hyperparameters} \begin{tabular}{lll} \toprule \textbf{Symbol} & \textbf{Meaning} & \textbf{Value} \\ \midrule $M$ & Number of oversampled candidates in FS-FPS & $100$ \\ $K$ & Number of selected starting points in FS-FPS & $8$ \\ $\theta_0$ & Cold-start initialization for SPB evaluators & all-ones vector \\ $\tau$ & Threshold multiplier for SPB lost-rate computation & $3.0$ \\ \bottomrule \end{tabular} \end{table}

\section{LSR-Synth Dataset}
\label{data}
The LSR-Synth dataset is designed to evaluate the true scientific equation discovery capability of large language model (LLM) based SR systems beyond memorization. Unlike benchmarks derived from well-known textbook equations, LSR-Synth introduces discovery-driven problems by combining established scientific terms with carefully constructed synthetic terms that are novel yet physically plausible.

\subsubsection{Scientific Domains}

LSR-Synth spans four scientific domains: chemistry, biology, physics, and material science. Chemistry problems focus on reaction kinetics and nonlinear concentration--time dynamics. Biology tasks emphasize population dynamics and growth processes with nonlinear regulatory effects. Physics equations are primarily drawn from dynamical systems, including oscillators, damping processes, and time-dependent forcing. Material science tasks cover stress--strain relationships, constitutive laws, and nonlinear material responses.

Each domain is selected to reflect realistic scientific modeling scenarios while maintaining diversity in equation structure and operator usage. The final LSR-Synth dataset contains 128 problems across these four domains.

\subsubsection{Equation Construction Pipeline}

The construction of LSR-Synth follows a multi-stage pipeline designed to ensure novelty, solvability, and scientific plausibility:

\paragraph{(1) Scientific Problem Selection.}
Representative scientific problems are first selected within each domain (e.g., reaction kinetics in chemistry or population growth in biology), along with a corresponding set of variables and physical interpretations.

\paragraph{(2) Known Term Generation.}
Given the problem description, a large language model is prompted to generate a set of commonly used and well-established mathematical terms that typically appear in standard formulations of the underlying scientific model.

\paragraph{(3) Synthetic Term Generation.}
In parallel, the LLM is prompted to generate synthetic terms that are novel in the given scientific context. These terms are designed to be mathematically valid and physically interpretable but are not part of standard textbook equations. Examples include higher-order nonlinearities, saturation effects, or unconventional time-dependent modulations.

\paragraph{(4) Equation Assembly and Solvability Check.}
Known terms and synthetic terms are combined into complete mathematical expressions. Each candidate equation is then verified for analytical or numerical solvability using numerical solvers (e.g., ODE solvers). Equations that are ill-posed or numerically unstable are discarded.

\paragraph{(5) Novelty Verification.}
To prevent trivial rediscovery or memorization, each constructed equation is evaluated for novelty using an LLM-based evaluator, which determines whether the equation requires data-driven reasoning rather than direct recall of known scientific laws.

\paragraph{(6) Data Generation.}
For equations that pass solvability and novelty checks, synthetic datasets are generated by numerically evaluating the equations under predefined parameter ranges and initial conditions. Both in-domain and out-of-domain (OOD) test sets are created when applicable to assess generalization.

\paragraph{(7) Expert Validation.}
Finally, the equations and their generated data are manually reviewed by subject matter experts to ensure scientific plausibility and consistency with the stated problem context.

\subsubsection{Symbolic Accuracy (SA)}

Symbolic Accuracy (SA) is the primary metric used to evaluate whether a discovered equation is symbolically equivalent to the ground-truth equation in LSR-Synth. Unlike traditional SR recovery metrics that rely on exact string matching, SA focuses on semantic equivalence.

Specifically, SA is computed using an LLM-based evaluator that compares the predicted equation and the ground-truth equation after removing numerical constants and fitted parameters. The evaluator determines whether the two expressions represent the same underlying mathematical relationship up to algebraic equivalence and parameterization.

This evaluation protocol accounts for variations in expression form (e.g., reordering of terms or alternative but equivalent formulations) and is particularly well-suited for LLM-based equation discovery systems that may output equations in diverse symbolic representations. The reliability of SA is validated through comparison with human expert judgments, showing high agreement.

\section{Worked Example for the Separable Form}
\label{app:separable_example}

This appendix provides a concrete example of the separable formulation used in
Section~\ref{sec:method}. The goal is to clarify the meanings and dimensions of
\(X\), \(y\), \(c_f\), \(\Phi_f\), \(\alpha\), and \(\beta\).

Consider a one-dimensional regression problem with \(n=100\) data points:
\[
\mathcal{D}=\{(x_i,y_i)\}_{i=1}^{100},
\qquad
x_i\in\mathbb{R},\quad y_i\in\mathbb{R}.
\]
The corresponding data matrix and target vector are
\[
X =
\begin{bmatrix}
x_1\\
x_2\\
\vdots\\
x_{100}
\end{bmatrix}
\in \mathbb{R}^{100\times 1},
\qquad
y =
\begin{bmatrix}
y_1\\
y_2\\
\vdots\\
y_{100}
\end{bmatrix}
\in \mathbb{R}^{100}.
\]

Now consider the candidate structure
\[
f(x;\theta)=a\sin(bx+c)+d,
\]
where the continuous parameter vector is
\[
\theta=(a,d,b,c),
\qquad
d_f=4.
\]
This structure contains two conditionally linear parameters and two nonlinear
parameters. We partition the parameters as
\[
\alpha =
\begin{bmatrix}
a\\
d
\end{bmatrix}
\in\mathbb{R}^{d_\alpha},
\qquad
d_\alpha=2,
\]
and
\[
\beta =
\begin{bmatrix}
b\\
c
\end{bmatrix}
\in\mathbb{R}^{d_\beta},
\qquad
d_\beta=2.
\]
Thus, \(d_f=d_\alpha+d_\beta=4\).

For a fixed nonlinear parameter vector \(\beta=(b,c)\), the prediction over all
100 data points can be written as
\[
f(X;\alpha,\beta)
=
\begin{bmatrix}
a\sin(bx_1+c)+d\\
a\sin(bx_2+c)+d\\
\vdots\\
a\sin(bx_{100}+c)+d
\end{bmatrix}.
\]
This vector admits the separable form
\[
f(X;\alpha,\beta)
=
c_f(X;\beta)+\Phi_f(X;\beta)\alpha,
\]
where
\[
c_f(X;\beta)=
\begin{bmatrix}
0\\
0\\
\vdots\\
0
\end{bmatrix}
\in\mathbb{R}^{100},
\]
and
\[
\Phi_f(X;\beta)=
\begin{bmatrix}
\sin(bx_1+c) & 1\\
\sin(bx_2+c) & 1\\
\vdots & \vdots\\
\sin(bx_{100}+c) & 1
\end{bmatrix}
\in\mathbb{R}^{100\times 2}.
\]
Therefore,
\[
\Phi_f(X;\beta)\alpha
=
\begin{bmatrix}
\sin(bx_1+c) & 1\\
\sin(bx_2+c) & 1\\
\vdots & \vdots\\
\sin(bx_{100}+c) & 1
\end{bmatrix}
\begin{bmatrix}
a\\
d
\end{bmatrix}
=
\begin{bmatrix}
a\sin(bx_1+c)+d\\
a\sin(bx_2+c)+d\\
\vdots\\
a\sin(bx_{100}+c)+d
\end{bmatrix}.
\]

This example illustrates the key point of the separable formulation: although
\(\Phi_f(X;\beta)\) depends on the nonlinear parameters \(\beta=(b,c)\), it does
not depend on the conditionally linear parameters \(\alpha=(a,d)\). Consequently,
once \(\beta\) is fixed, the optimal \(\alpha\) can be obtained by a linear
least-squares solve:
\[
\alpha^\star(\beta)
=
\left(\Phi_f(X;\beta)^\top\Phi_f(X;\beta)\right)^\dagger
\Phi_f(X;\beta)^\top
\left(y-c_f(X;\beta)\right).
\]
The original four-dimensional fitting problem over
\(\theta=(a,d,b,c)\) is therefore reduced to a two-dimensional projected problem
over only the nonlinear parameters \(\beta=(b,c)\):
\[
\min_{\beta\in B_f}
L_{\mathrm{proj}}(\beta)
=
\frac{1}{100}
\left\|
y-c_f(X;\beta)-\Phi_f(X;\beta)\alpha^\star(\beta)
\right\|_2^2.
\]

More generally, for \(n\) data points, input dimension \(d_x\), \(d_\alpha\)
conditionally linear parameters, and \(d_\beta\) nonlinear parameters, the
dimensions are
\[
X\in\mathbb{R}^{n\times d_x},
\qquad
y\in\mathbb{R}^{n},
\qquad
\alpha\in\mathbb{R}^{d_\alpha},
\qquad
\beta\in\mathbb{R}^{d_\beta},
\]
\[
c_f(X;\beta)\in\mathbb{R}^{n},
\qquad
\Phi_f(X;\beta)\in\mathbb{R}^{n\times d_\alpha},
\qquad
d_f=d_\alpha+d_\beta.
\]

\section{Detailed LLM-SR Trajectory Diagnostics}
\label{app:trajectory_diagnostics}
\label{app:lost_rate}

We conduct a fine-grained analysis of the LLM-SR search trajectories
to examine how inner-loop evaluation fidelity affects the discovery
process. The backbone LLM, prompt templates, random seeds, search
configuration, and computational budget are fixed between LLM-SR and
LLM-SR+, so their trajectory differences arise from the evaluator
feedback.

\subsection{Task-Level Improvement and Aggregate Results}

\paragraph{Task-level log-ratio improvement.}
For each task \(i\), we quantify the final NMSE improvement obtained
by replacing the original evaluator with SAGE-Fit as
\begin{equation}
    \rho_i
    =
    \log_{10}
    \left(
        \frac{
            \mathrm{NMSE}_{\mathrm{LLM\text{-}SR},i}
        }{
            \mathrm{NMSE}_{\mathrm{LLM\text{-}SR+},i}
        }
    \right).
    \label{eq:llmsr_log_ratio}
\end{equation}
A positive \(\rho_i\) indicates a lower final NMSE with SAGE-Fit,
while its magnitude measures the reduction in orders of magnitude.
For example, \(\rho_i=2\) corresponds to a \(100\)-fold reduction in
NMSE. We report the mean and median of \(\rho_i\) within each domain.

\begin{table}[H]
    \centering
    \small
    \setlength{\tabcolsep}{5pt}
    \renewcommand{\arraystretch}{1.05}
    \caption{
    Fine-grained diagnostics on LLM-SR trajectories. The overall mean
    log-ratio and average lost rate are weighted by the numbers of tasks
    in Biology, Chemistry, Physics, and Material Science
    (\(24\), \(36\), \(44\), and \(25\), respectively).}
    \label{tab:trajectory_diagnostics}
    \begin{tabular}{lcccc}
        \toprule
        & \multicolumn{2}{c}{Log-Ratio}
        & \multicolumn{2}{c}{Lost Rate (\%)} \\
        \cmidrule(lr){2-3}
        \cmidrule(lr){4-5}
        Domain
        & Mean
        & Median
        & Avg.
        & Max \\
        \midrule
        Biology
        & 2.52
        & 1.91
        & 24.06
        & 61.79 \\
        Chemistry
        & 4.09
        & 1.59
        & 22.35
        & 44.35 \\
        Physics
        & 1.39
        & 0.69
        & 6.39
        & 20.79 \\
        Material Science
        & 5.84
        & 5.67
        & 47.81
        & 65.75 \\
        \midrule
        Overall
        & 3.22
        & --
        & 22.16
        & 65.75 \\
        \bottomrule
    \end{tabular}
\end{table}

Table~\ref{tab:trajectory_diagnostics} presents two
complementary aspects of evaluator fidelity. First, the median
log-ratio is positive across all domains, indicating that a typical
task benefits from replacing the original evaluator with SAGE-Fit.
The improvement is strongest in Material Science, where the median
log-ratio reaches \(5.67\), corresponding to an NMSE reduction of
more than five orders of magnitude. Biology and Chemistry also show
substantial median improvements of \(1.91\) and \(1.59\),
respectively. Physics exhibits a smaller but consistently positive
median improvement of \(0.69\), suggesting that structure proposal
may account for a larger share of the remaining difficulty in its
highly oscillatory tasks.

Second, the refit-based lost rate measures how often inaccurate
fitting changes search decisions. Material Science has the highest
average lost rate at \(47.81\%\), followed by Biology at \(24.06\%\)
and Chemistry at \(22.35\%\). Physics is comparatively more robust,
but its most severe trajectory still loses \(20.79\%\) of the
discarded candidates. Across all \(129\) tasks, SAGE-Fit reduces the
final NMSE by an average of \(3.22\) orders of magnitude, while
\(22.16\%\) of discarded candidates on average outperform their
corresponding incumbents after refitting; the maximum reaches
\(65.75\%\). These results directly support the ``Good Structure,
Bad Score'' hypothesis: inaccurate parameter fitting can eliminate
promising expressions before they influence subsequent search.

\subsection{Illustrative Example of the Refit-Based Lost Rate}

We provide a concrete example to clarify the trajectory-level
diagnostic used to compute the refit-based lost rate. Consider a
baseline LLM-SR run that produces a chronological sequence of
candidate structures
\[
E=(f_1,f_2,\ldots,f_R),
\]
where \(R\) denotes the number of generated candidates. Among these
candidates, the original evaluator updates the incumbent solution only
at a few iterations. We denote these best-so-far updates by
\[
U=(u_1,u_2,\ldots,u_J).
\]
Here, \(u_j\) is the candidate that strictly improves the best-so-far
fitted error under the original evaluator.

For the interval after \(u_j\) and before \(u_{j+1}\), let \(C_j\)
denote the set of candidates generated in this interval but discarded
by the original evaluator. For the final update, the interval extends
from \(u_J\) to the end of the trajectory. We then refit both \(u_j\)
and every \(c\in C_j\) with SAGE-Fit. A discarded candidate \(c\) is
counted as lost if
\[
\widehat{S}_{\mathrm{SAGE}}(c)
<
\widehat{S}_{\mathrm{SAGE}}(u_j).
\]
This condition means that \(c\) would have achieved a lower fitted
error than the incumbent at that point if both candidates had been
evaluated by SAGE-Fit.

The refit-based lost rate is computed as
\[
\mathrm{LostRate}
=
\frac{
\sum_j \sum_{c\in C_j}
\mathbb{I}
\left[
\widehat{S}_{\mathrm{SAGE}}(c)
<
\widehat{S}_{\mathrm{SAGE}}(u_j)
\right]
}{
N_{\mathrm{compared}}
},
\]
where \(N_{\mathrm{compared}}\) denotes the number of discarded
candidates for which both the candidate and the corresponding
incumbent can be successfully refitted. A high lost rate indicates
that the original evaluator frequently assigns poor scores to
candidates that become competitive under more faithful parameter
fitting.

\begin{figure}[H]
    \centering
    \includegraphics[width=0.75\linewidth]
    {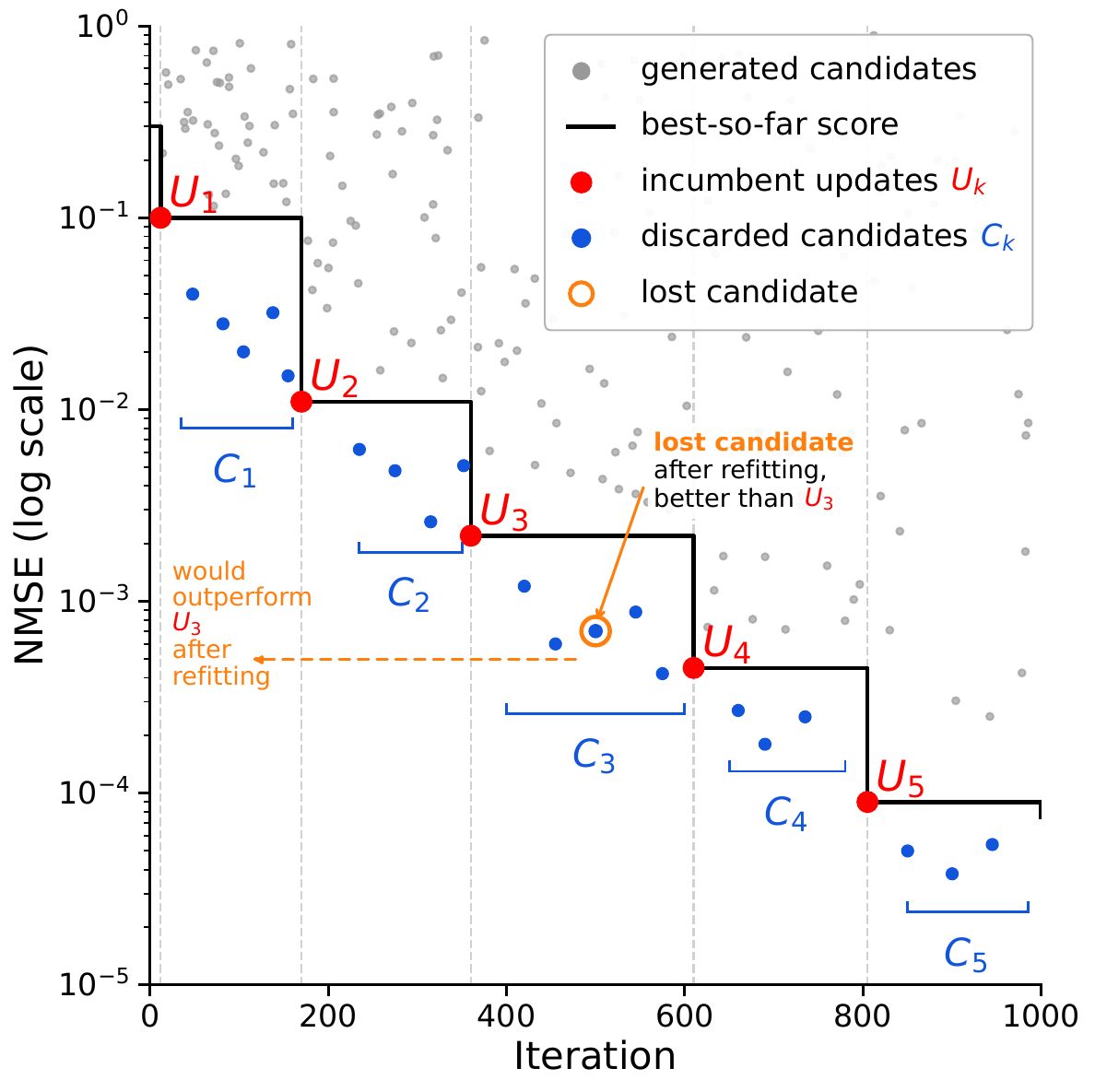}
    \caption{
    Schematic illustration of the refit-based lost rate. Gray points
    denote candidate structures generated along a baseline LLM-SR
    trajectory. The black step curve denotes the best-so-far fitted
    error under the original evaluator. Red points \(u_j\) indicate
    incumbent updates. Blue points denote discarded candidates
    \(c\in C_j\) between two consecutive incumbent updates. A
    discarded candidate is counted as lost if, after refitting with
    SAGE-Fit, it obtains a lower fitted error than the corresponding
    incumbent \(u_j\).}
    \label{fig:lost_rate_example}
\end{figure}

\section{Structure-Progress Bank (SPB) Design}
\label{app:spb_design}

The Structure-Progress Bank (SPB) is designed to analyze parameter-optimization failures in symbolic regression at the candidate level. It contains two complementary components. The first component consists of ground-truth equations together with representative parameterized structures associated with these equations. This component is used to illustrate how scientifically meaningful targets can induce substantially different fitting difficulties under different parameterizations. The second component consists of progress-associated candidates extracted from actual SR search trajectories. This component is used for the optimizer comparison in Section~\ref{sec:spb}. Together, these two components allow SPB to characterize both ground-truth-related parameterization patterns and fitting failures encountered during real discovery processes.

\subsection{Design Principles}

SPB is constructed according to three principles.

\paragraph{Diversity in parameterization complexity.}
The selected structures cover a broad range of parameterization patterns, from nearly linear coefficient fitting to nonlinear parameter placement. Parameters may appear as amplitudes, offsets, denominators, exponential rates, trigonometric frequencies, or exponents. This diversity ensures that SPB does not merely evaluate ordinary coefficient regression, but also probes nonlinear fitting regimes that frequently arise in SR.

\paragraph{Real discovery context.}
The progress-associated component of SPB is extracted from actual SR search trajectories rather than sampled from a synthetic expression grammar. Each candidate was generated by an outer-loop search process and therefore reflects the distribution of structures that SR systems actually consider during discovery, including near-miss candidates, redundant variants, and candidates whose scores are sensitive to parameter-fitting quality.

\paragraph{Empirical reference for fitting quality.}
For the progress-associated candidates used in Section~6.4, we evaluate each candidate using all compared optimizers and define its empirical reference score as the best NMSE achieved among them. This reference is not a global oracle and does not assume access to the true optimum. It is used only to quantify relative fitting failures under a fixed comparison set.

\subsection{Construction Protocol}

SPB is constructed from LSR-Synth problems and their corresponding SR search trajectories.

\paragraph{Ground-truth-related parameterized structures.}
For selected LSR-Synth problems, we retain the ground-truth equation together with representative parameterized candidate forms associated with the same scientific target. This component is used to inspect how different parameter placements change the fitting problem while remaining tied to a known ground-truth equation.

\paragraph{Progress-associated candidates.}
For the candidate-level optimizer comparison, we collect 200 candidates from 20 LSR-Synth problems, with 10 candidates selected from each problem. A candidate is selected if it is associated with observable progress in the search trajectory or if it becomes competitive after offline refitting. This selection strategy concentrates the diagnostic set on candidates for which evaluator fidelity can plausibly affect the subsequent search direction.

All candidates are required to be syntactically valid and numerically evaluable on the corresponding input domain. Candidates that produce invalid values, such as NaNs, overflows, or domain errors, are removed. Each remaining candidate is represented as a callable expression with free continuous parameters and is evaluated under the same initialization and fitting budget across optimizers.

For each progress-associated candidate $f_i$, the empirical reference score is defined as
\[
    \widehat{\mathcal{S}}_{\mathrm{ref}}(f_i)
    =
    \min_{e\in\mathcal{E}}
    \widehat{\mathcal{S}}_{e}(f_i),
\]
where
\[
\mathcal{E}
=
\{
\text{BFGS single},
\text{BFGS multi},
\text{SAGE-Fit w/o FS-FPS},
\text{SAGE-Fit}
\}.
\]
This definition is consistent with Section~6.4: SPB does not assume access to the globally optimal fitting score, but uses the best achieved score among the evaluated procedures as an empirical reference.

\subsection{SPB Statistics}

The progress-associated component used in the optimizer comparison contains 200 candidate structures from 20 LSR-Synth problems. The number of free parameters ranges from 2 to 11, with a median of 7. Each candidate is annotated with its source problem, generation event, total parameter count, nonlinear parameter count after VarPro reduction, fitted NMSE under each evaluator, and empirical reference NMSE.

In addition, SPB retains representative ground-truth-related parameterized structures for qualitative inspection. These examples are not used to define the denominator of the SPB lost rate in Section~\ref{sec:spb}. Instead, they explain why the same scientific target can induce substantially different fitting landscapes depending on how its parameters enter the expression.

\subsection{Representative Ground-Truth Structures and Associated Candidate Forms}
\label{app:spb_examples}

To illustrate the two components of SPB, we present representative examples from two task families: population growth with predation and chemical kinetics. For each case, we show the ground-truth equation and several parameterized candidate forms associated with the same problem. These examples are intended to demonstrate the diversity of parameter placements encountered in SPB, rather than to claim that every listed candidate is algebraically equivalent to the ground-truth equation.

\subsubsection{Population Growth with Predation}

The ground-truth equation is
\[
\frac{dP}{dt}
=
0.954\, P\left(1-\frac{P}{96.9}\right)
+
0.954\, P^{0.333}.
\]
This equation combines logistic growth with an allometric term. The associated candidate structures in SPB instantiate this target with different parameterization patterns, including time-dependent modulation, exponential relaxation, and trigonometric forcing. These variants illustrate how the same scientific problem can induce fitting objectives of substantially different difficulty.

\begin{table}[H]
\centering
\caption{
Representative parameterized candidate forms for the population-growth task. The table shows how candidate structures associated with the same ground-truth problem can vary in parameter count and nonlinear parameter placement.
}
\label{tab:structure1_variants}
\small
\begin{tabular}{p{0.12\textwidth}p{0.15\textwidth}p{0.63\textwidth}}
\toprule
\textbf{Variant} & \textbf{\# Params} & \textbf{Candidate Form} \\
\midrule
Variant 1 & 2 &
$\displaystyle \frac{dP}{dt} = a_1 P (1 - P/a_2)$ \\
& & Logistic-growth form with low parameter complexity. \\
\midrule
Variant 2 & 5 &
$\displaystyle \frac{dP}{dt} = a_1 P' (1 - P'/a_2) a_5,\quad
P' = (P - a_3)e^{-(t-a_4)} + a_3$ \\
& & Time-dependent adjustment with exponential relaxation. \\
\midrule
Variant 3 & 8 &
$\displaystyle
\frac{dP}{dt}
=
a_1 P(1-P/a_2)e^{-a_3t}e^{-t/a_4}
(1+a_5\sin(2\pi a_6t))
+a_7+a_8$ \\
& & Exponential rates and sinusoidal frequency introduce nonlinear fitting difficulty. \\
\midrule
Variant 4 & 10 &
$\displaystyle
\frac{dP}{dt}
=
a_1P(1-P/a_2)+a_3t+a_4+a_5\sin(a_6t)
+a_7P+a_8t^2+a_9\exp(-a_{10}t)$ \\
& & Mixed linear, polynomial, sinusoidal, and exponential components. \\
\bottomrule
\end{tabular}
\end{table}

\subsubsection{Chemical Kinetics with Michaelis--Menten-like Dynamics}

The ground-truth equation is
\[
\frac{dA}{dt}
=
-0.679 A^2
-
0.679 A
+
0.679 \sin(\log(A+1)).
\]
This equation combines second-order decay with a logarithmic-sinusoidal term. The associated candidate structures include polynomial decay, rational reaction terms, and Michaelis--Menten-like denominators. These forms create nonlinear least-squares problems whenever parameters alter the basis itself rather than merely rescaling fixed basis functions.

\begin{table}[H]
\centering
\caption{
Representative parameterized candidate forms for the chemical-kinetics task. The table highlights how denominator parameters and additional reaction terms can create nonlinear fitting objectives.
}
\label{tab:structure29_variants}
\small
\begin{tabular}{p{0.12\textwidth}p{0.15\textwidth}p{0.63\textwidth}}
\toprule
\textbf{Variant} & \textbf{\# Params} & \textbf{Candidate Form} \\
\midrule
Variant 1 & 3 &
$\displaystyle \frac{dA}{dt} = -a_1A^2 - \frac{a_2A}{a_3+A}$ \\
& & Michaelis--Menten-like denominator parameterization. \\
\midrule
Variant 2 & 3 &
$\displaystyle \frac{dA}{dt} = -a_1A - a_2A^2 - a_3tA$ \\
& & Time-dependent first-order decay with linearly separable coefficients. \\
\midrule
Variant 3 & 5 &
$\displaystyle \frac{dA}{dt} = -a_1A^2 - \frac{a_2A}{a_3+A} - a_4A + a_5$ \\
& & Extended reaction form with both denominator and additive terms. \\
\midrule
Variant 4 & 2 &
$\displaystyle \frac{dA}{dt} = -a_1A + \sum_i a_i A^{i+2}$ \\
& & Power-series form with coefficient-style parameters. \\
\bottomrule
\end{tabular}
\end{table}

\subsection{Representative SPB Rescue Cases}
\label{app:spb_rescue_cases}

Table~\ref{tab:spb_rescue} reports representative candidates for which the full SAGE-Fit evaluator achieves a substantially lower NMSE than its ablated variant without function-space farthest-point sampling. The failure ratio is computed as
\[
    \frac{
    \widehat{\mathcal{S}}_{\text{w/o FS-FPS}}(f_i)
    }{
    \widehat{\mathcal{S}}_{\text{SAGE-Fit}}(f_i)
    }.
\]
Both evaluators use the same number of final nonlinear solver calls. Therefore, these cases isolate the effect of selecting starting points in function space rather than relying on random multi-start initialization in parameter space.

\begin{table}[H]
    \centering
    \caption{
    Representative candidates rescued by SAGE-Fit relative to SAGE-Fit w/o FS-FPS.
    The failure ratio is computed as
    $\widehat{\mathcal{S}}_{\text{w/o FS-FPS}}(f_i)/
    \widehat{\mathcal{S}}_{\text{SAGE-Fit}}(f_i)$.
    }
    \label{tab:spb_rescue}
    \begin{tabular}{lcccc}
        \toprule
        \textbf{Problem}
        & \textbf{Event}
        & \textbf{\# Params}
        & \textbf{SAGE-Fit NMSE}
        & \textbf{Failure Ratio} \\
        \midrule
        structure13 & 9 & 8 & $3.96 \times 10^{-6}$ & $6.38 \times 10^{4}$ \\
        structure14 & 9 & 6 & $1.83 \times 10^{-4}$ & $5.47 \times 10^{2}$ \\
        structure15 & 5 & 8 & $4.11 \times 10^{-4}$ & $2.43 \times 10^{2}$ \\
        structure20 & 7 & 6 & $8.68 \times 10^{-4}$ & $1.15 \times 10^{2}$ \\
        \bottomrule
    \end{tabular}
\end{table}

These rescue cases indicate that FS-FPS changes the basins reached by the final local refinements. Although random multi-start initialization and FS-FPS use the same number of final solver calls, random starts can still concentrate on redundant regions of the landscape. By selecting behaviorally diverse initializations, SAGE-Fit improves the reliability of candidate-level evaluation under the same refinement budget.

\end{document}